\documentclass[conference]{IEEEtran}
\IEEEoverridecommandlockouts

\usepackage{cite}
\usepackage{amsmath,amssymb,amsfonts}
\usepackage{graphicx}
\usepackage{textcomp}
\def\BibTeX{{\rm B\kern-.05em{\sc i\kern-.025em b}\kern-.08em
    T\kern-.1667em\lower.7ex\hbox{E}\kern-.125emX}}

\usepackage{boldline}
\usepackage{subfig}
\usepackage{bm}

\DeclareMathOperator*{\argmin}{arg\,min} 

\usepackage[noend]{algpseudocode}
\usepackage{algorithm}
\usepackage[hidelinks]{hyperref}
\usepackage[dvipsnames]{xcolor}

\begin{document}

\title{
    Federated Learning with \\ Intermediate Representation Regularization

    \thanks{This work was supported by the Institute of Information and Communications Technology Planning and Evaluation (IITP) Grant funded by the Korea Government (MSIT) (Artificial Intelligence Innovation Hub) under Grant 2021-0-02068 and in part by the National Research Foundation of Korea (NRF) grant funded by the Korea government (MSIT) (No. 2020R1A4A1018607) *Dr. CS Hong is the corresponding author.}
}


\author{\IEEEauthorblockN{Ye Lin Tun, Chu Myaet Thwal, Yu Min Park, Seong-Bae Park, Choong Seon Hong*}
\IEEEauthorblockA{\textit{Department of Computer Science and Engineering} \\
\textit{Kyung Hee University}\\
Yongin-si, Republic of Korea \\
\{yelintun, chumyaet, yumin0906, sbpark71, cshong\}@khu.ac.kr}
}


\maketitle

\begin{abstract}

In contrast to centralized model training that involves data collection, federated learning (FL) enables remote clients to collaboratively train a model without exposing their private data. However, model performance usually degrades in FL due to the heterogeneous data generated by clients of diverse characteristics. One promising strategy to maintain good performance is by limiting the local training from drifting far away from the global model. Previous studies accomplish this by regularizing the distance between the representations learned by the local and global models. However, they only consider representations from the early layers of a model or the layer preceding the output layer. In this study, we introduce FedIntR, which provides a more fine-grained regularization by integrating the representations of intermediate layers into the local training process. Specifically, FedIntR computes a regularization term that encourages the closeness between the intermediate layer representations of the local and global models. Additionally, FedIntR automatically determines the contribution of each layer's representation to the regularization term based on the similarity between local and global representations. We conduct extensive experiments on various datasets to show that FedIntR can achieve equivalent or higher performance compared to the state-of-the-art approaches. Our code is available at \textcolor{TealBlue}{\url{https://github.com/YLTun/FedIntR}}.


\end{abstract}

\vspace{2mm}

\section{Introduction}
The real world data essential for powering intelligent services is spread across numerous edge devices (e.g., IoT devices, personal smartphones, or data silos of different organizations). Although deep learning can benefit from large datasets produced by mass collection, rising security concerns and privacy regulations~\cite{10.5555/3152676} may prohibit a server from acquiring the data from edge devices. Such limitations may render the centralized training of deep neural network models infeasible. Federated learning (FL)~\cite{mcmahan2017communication, kairouz2021advances, Li2020FederatedLC}, which allows edge devices to collaboratively train a model without sharing their data with a central server, has come to light as a viable option to meet these requirements. In particular, the FedAvg~\cite{mcmahan2017communication} algorithm has emerged as the de facto approach for model training in distributed environments with data privacy concerns. FedAvg operates by having each edge device train a local model on its own data before sending the trained parameters to the server. The server aggregates the received parameters into a single global model that inherits the trained capabilities of the local models.

\begin{figure}[tbp]
    \centering
    \includegraphics[width=\linewidth]{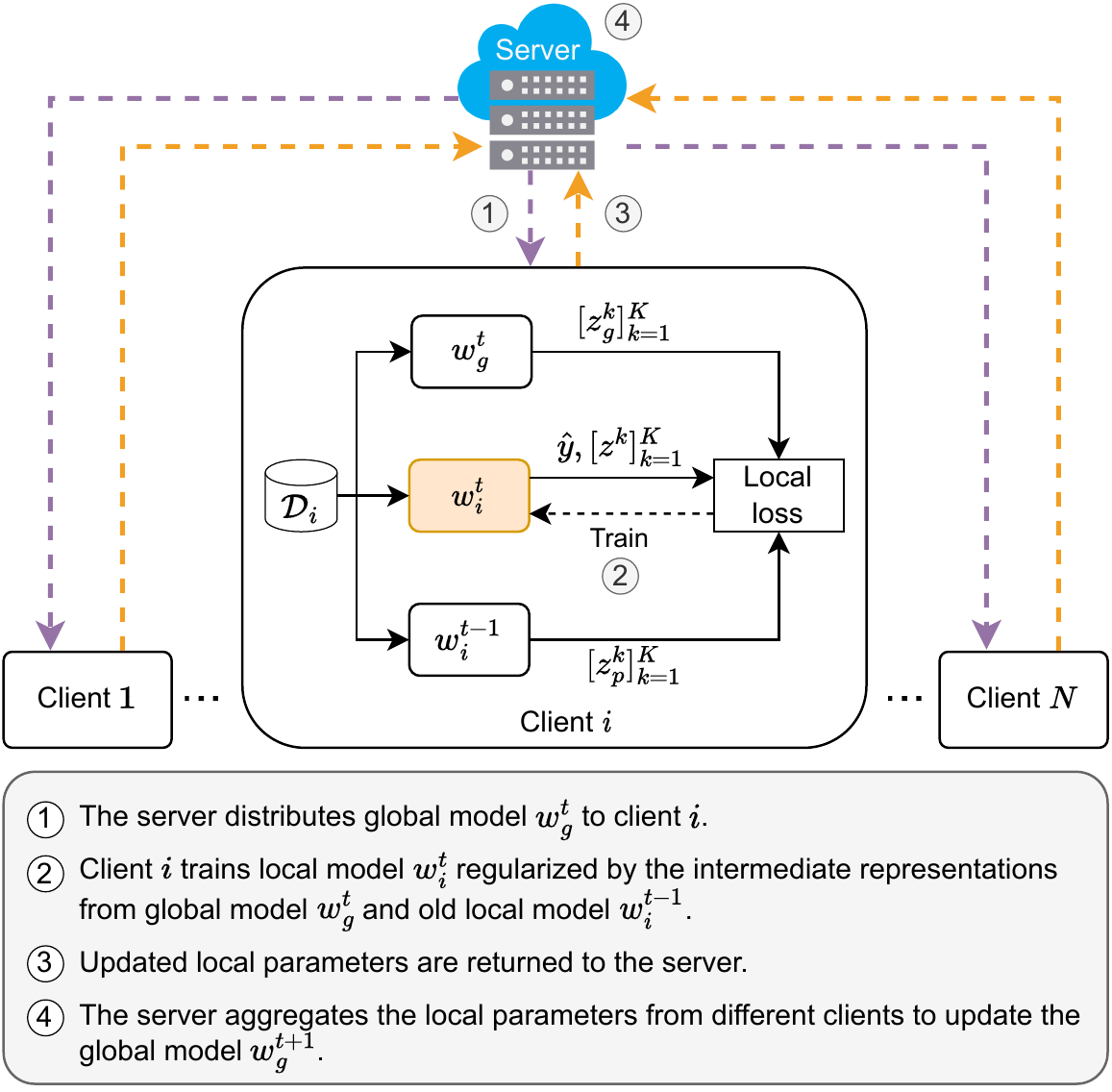}
    \caption{An Overview of federated learning with FedIntR. FedIntR incorporates regularization into the local training step (i.e., step2) of FedAvg.}
    \label{fig:federated_learning}
\end{figure}


\begin{figure*}[tb]
    \centering
    \includegraphics[width=\textwidth]{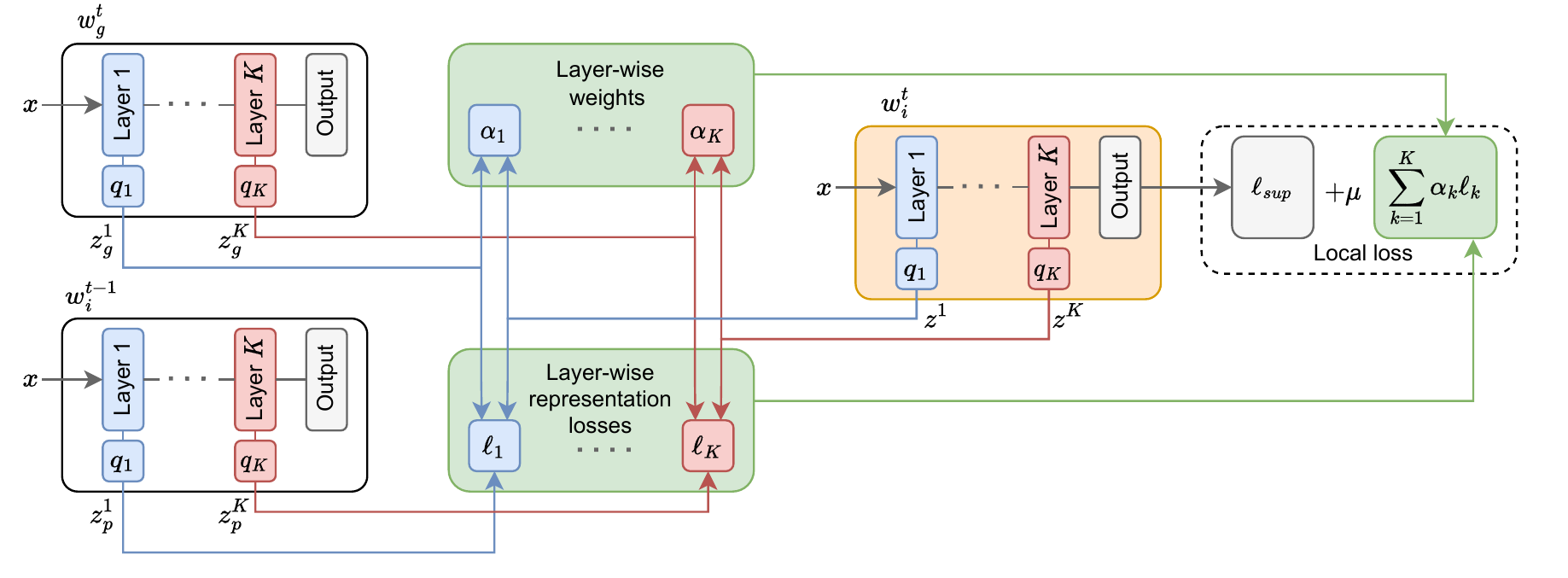}
    \caption{Local loss computation in FedIntR. Using the intermediate representations $z$ from the local model $w_i^t$, global model $w_g^t$ and old local model $w_i^{t-1}$, FedIntR computes a regularization term. Together, this regularization term and the cross entropy loss $\ell_{sup}$ make up the local training loss.
    }
    \label{fig:system_model}
\end{figure*}

\vspace{5mm}

In practice, large data heterogeneity may arise in an FL system since the local data on each device varies depending on the nature and behavior of the device. Heterogeneous data has been found to be a major issue that causes slow convergence and suboptimal model performance in federated learning \cite{li2020federated, zhao2018federated}. To address this, previous studies have incorporated various forms of regularization into the local training loss to reduce the divergence of local models from the global model \cite{li2020federated, karimireddy2020scaffold, li2021model, son2021compare}. The distance between local and global parameters can be limited using $\ell_2$-norm~\cite{li2020federated} or control variates~\cite{karimireddy2020scaffold}. A recent popular approach is to use representations such as MOON~\cite{li2021model} and FedCKA~\cite{son2021compare}, which adopt contrastive loss~\cite{chen2020simple, he2020momentum, misra2020self} to promote the closeness between local and global representations. However, these studies only focus on a few specific layers of the model to extract the representations. In particular, MOON extracts representations from the layer prior to the output layer, whereas FedCKA extracts from the first two layers based on the assumption that naturally similar layers are the most critical to improving the performance. Nevertheless, we argue that utilizing representations from all intermediate layers while assigning a proper weight to each of their contributions can offer a more fine-grained regularization to the training process.

Hence, we explore the idea of using representations extracted from intermediate layers to overcome the performance degradation caused by non-IID data in FL. We propose \mbox{\textbf{FedIntR}}, which promotes the similarity between \textbf{Int}ermediate layer \textbf{R}epresentations  the  of local and global models. Specifically, FedIntR computes a regularization term based on the contrastive loss~\cite{chen2020simple, li2021model} using local and global intermediate representations. Moreover, FedIntR automatically calculates layer-wise weights to determine each intermediate layer's contribution to the regularization term. Analogous to the idea of FedCKA~\cite{son2021compare}, where similar layers are more important for improving the performance, FedIntR assigns a larger contribution weight to the layers with a higher global and local representation similarity. To demonstrate the effectiveness of FedIntR, we conduct extensive experiments in comparison with various state-of-the-art FL approaches. Moreover, we evaluate FedIntR in different FL settings with varying degrees of data heterogeneity, numbers of clients, and local training epochs to ensure that it can maintain good performance in these scenarios.

\section{Background and Related Work}

\subsection{Federated Learning}


The FedAvg~\cite{mcmahan2017communication} algorithm is the first federated learning framework that enables collaborative model training while preserving the privacy of user devices. Fig.~\ref{fig:federated_learning} depicts the overview of a typical FL framework, which consists of four key steps at each training round $t$: (i) the central server distributes the global model $w_g^t$ to the clients; (ii) each client $i$ trains its local model $w_i^t$ on its private data $\mathcal{D}_i$; (iii) the trained local models are returned to the server; and (iv) the server aggregates the received local models into an updated global model $w_g^{t+1}$. FedAvg~\cite{mcmahan2017communication} also highlights the non-IID characteristics of an FL environment, and since then, improving the performance of an FL model under such constraints has been the subject of many studies.

\subsection{Tackling Data Heterogeneity}

Many FL studies rely on three distinct strategies for mitigating the performance degradation caused by heterogeneous data: (i) adding regularization to the local training, (ii) modifying the aggregation scheme, and (iii) fitting personalized models for the clients. We discuss each of these strategies as follows.

\vspace{1mm}
\noindent
\textbf{Local Training.} FedProx~\cite{li2020federated} includes a proximal term in the local training objective in order to restrict the Euclidean distance between the local and global models. MOON~\cite{li2021model} introduces model-contrastive loss which is inspired by the NT-Xent (Normalized Temperature-scaled Cross Entropy) loss \cite{sohn2016improved, wu2018unsupervised, oord2018representation} used in SimCLR~\cite{chen2020simple} for self-supervised representation learning. The NT-Xent loss is designed to minimize the distance between the representations of a positive pair (i.e., augmented views of the same sample) and maximize the distance between the representations of negative pairs (i.e., augmented views from different samples). Model-contrastive loss in MOON aims to minimize the distance between representations extracted by the global and local models while simultaneously maximizing the distance between representations from the current local model and the old local model. MOON is based on the idea that, since the global model has been trained on the entire dataset, its representations can effectively regularize the local training. SCAFFOLD~\cite{karimireddy2020scaffold} lowers the client-variance in the local updates by minimizing the difference between the local and global control variates. Similar to MOON, FedCKA~\cite{son2021compare} uses global and local model representations to regularize the local training and shows that using deeper layers for regularization can harm the performance. Centered kernel alignment (CKA)~\cite{kornblith2019similarity} is used in FedCKA to calculate the distance between the representations, whereas MOON employs cosine similarity.


\vspace{1mm}
\noindent
\textbf{Aggregation.} Another way to combat the downsides of data heterogeneity is to design a more intelligent aggregation scheme. PFNM~\cite{yurochkin2019bayesian} relies on Bayesian nonparametric methods to match the neurons of different client models before the aggregation process. However, PFNM only works with fully connected neural networks. Therefore, FedMA~\cite{wang2020federated} proposes to match in a layer-wise fashion, which is applicable for CNN and LSTM models as well. Inspired by the effectiveness of SGD momentum in dampening gradient oscillations, FedAvgM~\cite{hsu2019measuring} incorporates momentum in the aggregation process. FedNova~\cite{wang2020tackling} takes into account the varying number of local update steps completed by the clients and normalizes the local update by the number of steps prior to aggregation. pFedLA~\cite{ma2022layer} generates personalized models by using hypernetworks to determine the layer-wise contribution of each client model to the aggregation step.

\vspace{1mm}
\noindent
\textbf{Personalized Federated Learning.} Personalized models can improve the performance in non-IID settings by fitting the diverse distributions of the clients' data \cite{t2020personalized, fallah2020personalized, jiang2019improving, zhang2020personalized}. One simple and effective personalized FL mechanism is to group clients with similar characteristics into clusters, and cluster members can collectively train a personalized model. Many approaches~\cite{sattler2020clustered, duan2021fedgroup, xie2020multi, sattler2020byzantine, ghosh2019robust} cluster the clients based on the similarity of local updates received at the server. Others~\cite{ghosh2020efficient, mansour2020three, tun4148978contrastive} perform clustering on the client-side, where each client can choose their own personalized model from a set of available models based on performance.

\vspace{1mm}
Our proposed method, FedIntR, is one of the techniques that incorporates regularization into the local training process. A recent study~\cite{kaku2021intermediate} demonstrates that self-supervised learning with additional loss across intermediate layers can enhance the model's performance in the downstream tasks. Inspired by~\cite{kaku2021intermediate}, we explore the integration of intermediate representations into the FL process to enable a more effective regularization for data heterogeneity issues. In contrast to our approach, MOON~\cite{li2021model} only uses the representations from the layer preceding the output layer, and FedCKA~\cite{son2021compare} is based on the idea of minimizing the distance between similar layer representations and, therefore, only considers the early layers (the first two layers) of the model. FedIntR can also be considered a more general approach where we don’t have to manually define which layers to include in the regularization term since it automatically determines the contribution of different intermediate layers to the regularization based on the similarities between their local and global representations.




\begin{algorithm}[tbp]
\caption{FedIntR}
\label{alg:fedintr}
\begin{algorithmic}[1]

    \State \textbf{Input:}  number of training rounds $T$, number of local epochs $E$, number of clients $N$, number of intermediate layers $K$, temperature $\tau$, learning rate $\eta$, weighting parameter $\mu$
    \State \textbf{Output:} global model $w^T_{g}$
    \item[] 
    
    \State \textbf{Server executes:}
    \State initialize $w^0_{g}$
    \For{each round $t=0,1,\dots,T-1$}
        \For{each client $i=1,2,\dots,N$ in parallel}
            \State $w^t_i \leftarrow \text{LocalUpdate}(i, w^t_g)$
        \EndFor
        \State $w^{t+1}_g \leftarrow \sum^N_{i=1} \frac{|\mathcal{D}_i|}{|\mathcal{D}|} w^t_i$ 
    \EndFor
    \State return $w^T_g$
    \item[] 
    
    \State \textbf{Client executes:} LocalUpdate($i, w^t_g$):
    \State $w^{t-1}_i \leftarrow w^{t}_i$
    \State $w^{t}_i \leftarrow w^t_g$
    \For{each epoch $e=0,1,\dots,E-1$}
        \For{each batch $\{x, y\} \in \mathcal{D}_i$}
            \State $\hat{y}, [z^k]^K_{k=1} \leftarrow w^t_i(x)$
            \vspace{1mm}
            \State $[z^k_g]^K_{k=1} \leftarrow w^t_g(x)$
            \vspace{1mm}
            \State $[z^k_p]^K_{k=1} \leftarrow w^{t-1}_i(x)$
            \vspace{1mm}
            \State $\ell_{sup} \leftarrow CrossEntropyLoss(\hat{y}, y)$
            \For{$k=1,2,\dots,K$}
                \vspace{1mm}
                \State $\alpha_k \leftarrow  \frac{\exp(\text{sim}(z^k, z^k_{g})/\tau)}{\sum_{\hat{k}=1}^K \exp(\text{sim}(z^{\hat{k}}, z^{\hat{k}}_{g})/\tau)}$
                \State $\ell_k \leftarrow - \log \frac{\exp(\text{sim}(z^k, z^k_{g})/\tau)}{\exp(\text{sim}(z^k, z^k_{g})/\tau) + \exp(\text{sim}(z^k, z^k_{p})/\tau)}$
            \EndFor
            \State $\mathcal{L} \leftarrow \ell_{sup} + \mu \sum^K_{k=1} \alpha_k \ell_k$
            \State $w^t_i \leftarrow w^t_i - \eta \nabla \mathcal{L}$
        \EndFor
    \EndFor
    \State return $w^{t}_i$

\end{algorithmic}
\end{algorithm}

\section{Method}

\subsection{Federated Learning Objective}

The goal of a federated learning system is to train a global model $w_g$ by solving:
\begin{equation}
    w_g^* = \argmin_{w_g} \sum_{i=1}^N \frac{|\mathcal{D}_i|}{\mathcal{|D|}} \mathcal{L}_i (w_g),
    \label{eqn:fl_obj}
\end{equation}
where $\mathcal{L}_i$ and $\mathcal{D}_i$ represent the local loss and the dataset of the $i$-th client, $|\mathcal{D}|=\sum_{i=1}^N |\mathcal{D}_i|$ and $N$ is the total number of clients. From Fig.~\ref{fig:federated_learning}, we can observe that FedIntR only modifies the local training process of vanilla FedAvg, similar to MOON and FedCKA. In addition to the cross entropy loss, FedIntR adds a regularization term computed with the help of intermediate representations from the global model $w_g^t$ and the old local model $w_i^{t-1}$ from the previous round.


\subsection{Local Training Process of FedIntR}


Fig.~\ref{fig:system_model} illustrates how the local training loss is calculated in our FedIntR framework, while Algorithm~\ref{alg:fedintr} describes the whole procedure in detail. The local training process begins with the $i$-th client receiving the global model $w_g^t$ from the central server. The client synchronizes the old local model $w_i^{t-1}$ with the current one $w_i^t$ (which we want to train), followed by synchronizing $w_i^t$ with $w_g^t$. Suppose that the deployed model structure contains a total of $K$ layers capable of extracting representations (e.g., convolutional or fully connected layers). Given an input $x$, FedIntR extracts representations $z^k$, $z^k_p$, and $z^k_g$ from the $k$-th layer of the current local model $w_i^t$, old local model $w_i^{t-1}$, and global model $w_g^t$, respectively. (A representation is obtained by passing the layer output through a projection head $q_k$ as shown in Fig.~\ref{fig:system_model}.) Inspired by MOON, FedIntR computes the representation loss for $k$-th layer as
\begin{equation}
    \ell_k = - \log \frac{\exp(\text{sim}(z^k, z^k_{g})/\tau)}{\exp(\text{sim}(z^k, z^k_{g})/\tau) + \exp(\text{sim}(z^k, z^k_{p})/\tau)}
    \label{eqn:representation_loss},
\end{equation}
where $\tau$ is the temperature parameter and $\text{sim}(.,.)$ is the similarity function. We use the cosine similarity function, which is defined by
\begin{equation}
    \text{sim}(z_i, z_j)=\frac{z_i}{{\lVert z_i \rVert}_2} \cdot \frac{z_j}{{\lVert z_j \rVert}_2}
    \label{eqn:cosine_similarity}.
\end{equation}
The layer-wise representation loss $\ell_k$ encourages the local representation $z^k$ to be closer to global representation $z^k_g$ while moving further from $z^k_p$.

The layer-wise weight $\alpha_k$, representing the contribution of $\ell_k$ to the regularization term, is determined based on the similarity of $z^k$ and $z^k_g$ using the softmax function. Specifically, $\alpha_k$ is computed as follows:
\begin{equation}
    \alpha_k = \frac{\exp(\text{sim}(z^k, z^k_{g})/\tau)}{\sum_{\hat{k}=1}^K \exp(\text{sim}(z^{\hat{k}}, z^{\hat{k}}_{g})/\tau)}
    \label{eqn:contribution_weight},
\end{equation}
where $\sum_{k=1}^K \alpha_k=1$. Equation~\eqref{eqn:contribution_weight} assigns a higher weight value to the layer $k$ with a higher degree of similarity between $z^k$ and $z_g^k$.

FedIntR computes $\alpha_k$ and $\ell_k$ for all layers $k \in [1,2,\dots,K]$. After that, we can incorporate $[\alpha_k]_{k=1}^K$ and $[\ell_k]_{k=1}^K$ into the local training loss as the regularization term. The local loss $\mathcal{L}$ is defined as:
\begin{equation}
    \mathcal{L} = \ell_{sup} + \mu \sum^K_{k=1} \alpha_k \ell_k
    \label{eqn:local_loss},
\end{equation}
where $\ell_{sup}$ represents the cross-entropy loss, and the second component is the regularization term, with $\mu$ being the balancing parameter. 


\section{Experiments}

In this section, we discuss various experimental setups and evaluation results used to verify the effectiveness of FedIntR relative to other baseline approaches.

\subsection{Experimental Settings}

FedIntR is evaluated on four datasets: Fashion-MNIST~\cite{xiao2017/online}, SVHN~\cite{netzer2011reading}, CIFAR-10~\cite{krizhevsky2009learning}, and CIFAR-100~\cite{krizhevsky2009learning}, in comparison with various state-of-the-art FL approaches including FedAvg~\cite{mcmahan2017communication}, FedProx~\cite{li2020federated}, MOON~\cite{li2021model}, and FedCKA~\cite{son2021compare}. We use a small CNN model for the Fashion-MNIST, SVHN, and CIFAR-10 datasets, and ResNet-20~\cite{he2016deep, Idelbayev18a} for the CIFAR-100 dataset. Our CNN model is composed of three $3\times3$ convolutional layers with 8, 16, and 32 channels. Each convolutional layer is ReLU activated and followed by a $2\times2$ max pooling. The output of the final convolutional layer goes through two fully connected layers with 128 and 96 neurons. Both fully connected layers are also ReLU activated. FedIntR extracts representations from the three convolutional layers and the two fully connected layers of the CNN model. Similar to MOON, our projection head is a 2-layer MLP with an output dimension of 256. For the ResNet-20 model, the representations are retrieved from the first convolutional layer and each end of the three ResNet blocks.

We use the PyTorch~\cite{paszke2019pytorch} framework to implement FedIntR and other baseline methods. Referring to MOON, local training is performed using the SGD optimizer with a learning rate of 0.01, weight decay of 0.00001, and momentum of 0.9. We set the batch size as 512, the number of local epochs as 10, and the number of training rounds as 100. We use random horizontal flip as augmentation during the training except for the SVHN dataset. For both FedIntR and MOON, the temperature $\tau$ value is set to 0.5.

\begin{table}[tbp]
\centering
\caption{Test accuracy (\%) on 10 clients with $\beta=0.5$. We tune $\mu$ for each method, and the best $\mu$ value is shown in parentheses.}
\label{tab:mult_datasets}
\begin{tabular}{l|llll}
\hlineB{3}
        & \multicolumn{1}{c}{\begin{tabular}[c]{@{}c@{}}Fashion-\\ MNIST\end{tabular}} & \multicolumn{1}{c}{SVHN} & \multicolumn{1}{c}{CIFAR-10} & \multicolumn{1}{c}{CIFAR-100} \\ \hlineB{3}
FedAvg  & 88.90                 & 86.04                 & 65.82                 & 40.43                 \\
FedCKA  & 88.85 (3)             & 86.47 (1)             & 66.55 (0.1)           & \textbf{41.91} (10)   \\
FedProx & 88.95 (0.001)         & 86.42 (0.01)          & 65.03 (0.01)          & 39.91 (0.1)           \\
MOON    & \textbf{89.15} (1)    & 86.61 (10)            & 66.35 (5)             & 40.46 (0.1)           \\ \hline
FedIntR & \textbf{89.15} (10)   & \textbf{87.17} (10)   & \textbf{67.23} (3)    & 41.48 (10)            \\ \hlineB{3}
\end{tabular}
\end{table}

\begin{table}[tbp]
\centering
\caption{Test accuracy (\%) of FedIntR with different $\mu$ values.}
\label{tab:mu_vs_accuracy}
\begin{tabular}{c|cccc}
\hlineB{3}
$\mu$ & Fashion-MNIST  & SVHN           & CIFAR-10       & CIFAR-100      \\ \hlineB{3}
0.1     & 88.87          & 86.85          & 65.23          & 41.00           \\
1       & 89.07          & 85.89          & 65.66          & 40.43          \\
3       & 88.67          & 87.05          & \textbf{67.23} & 40.99          \\
5       & 89.13          & 87.01          & 66.99          & 41.21          \\
10      & \textbf{89.15} & \textbf{87.17} & 65.89          & \textbf{41.48} \\ \hlineB{3}
\end{tabular}
\end{table}

\begin{figure*}[tbp]
    \centering
    \subfloat[$\beta=5.0$]{\includegraphics[width=0.325\textwidth]{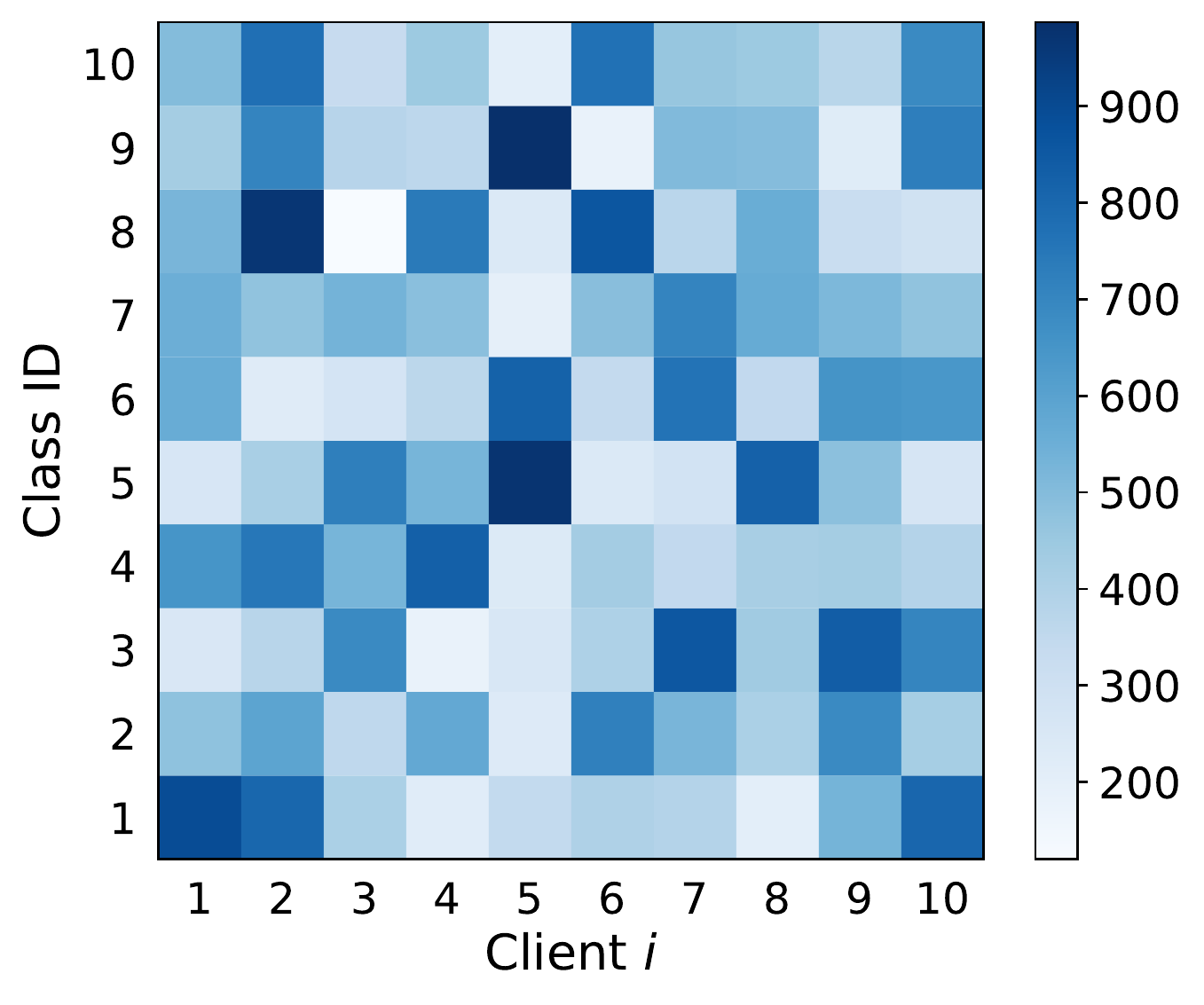} \label{subfig:beta_5}} 
	\subfloat[$\beta=0.5$]{\includegraphics[width=0.325\textwidth]{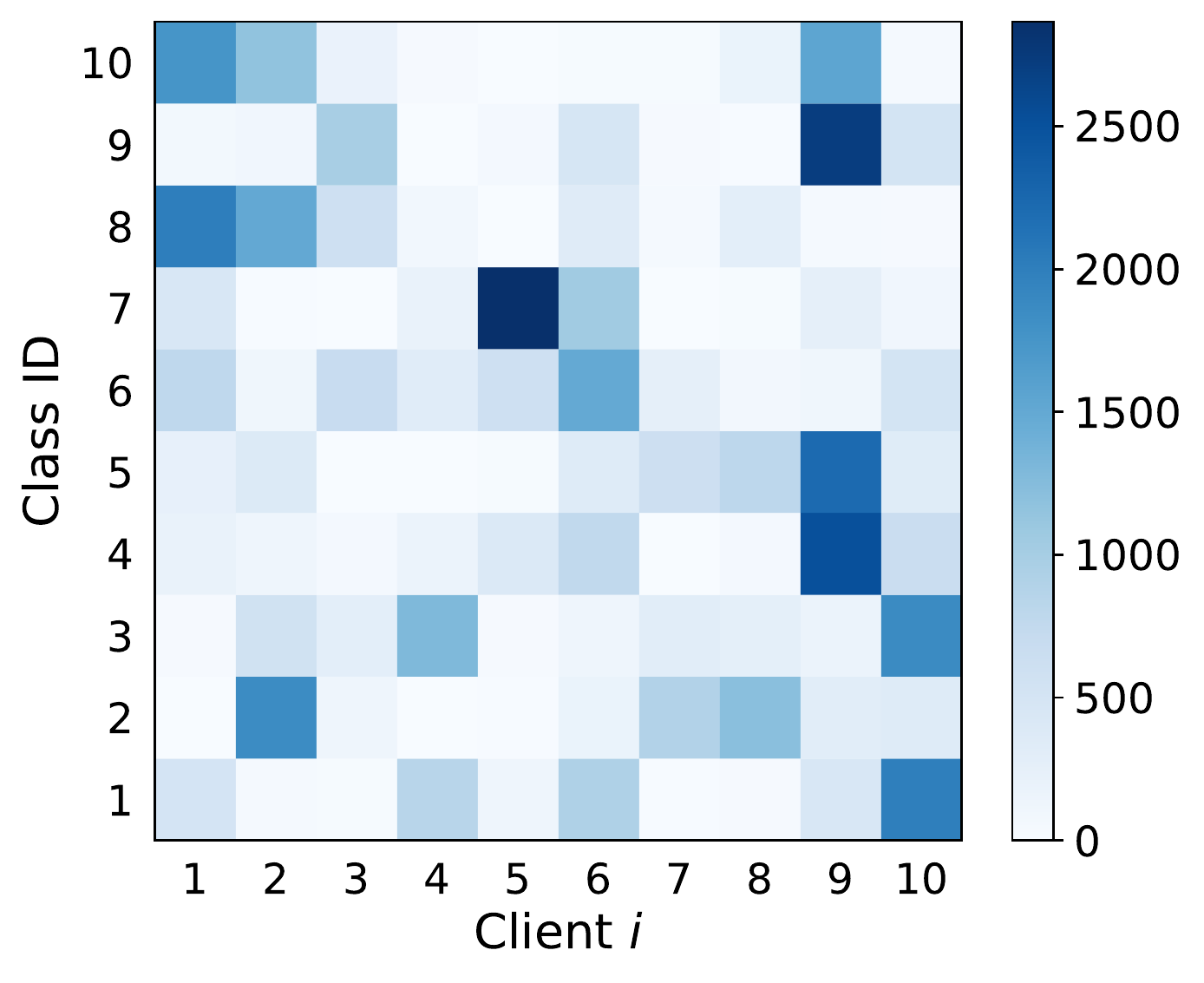} \label{subfig:beta_0.5}} 
	\subfloat[$\beta=0.1$]{\includegraphics[width=0.325\textwidth]{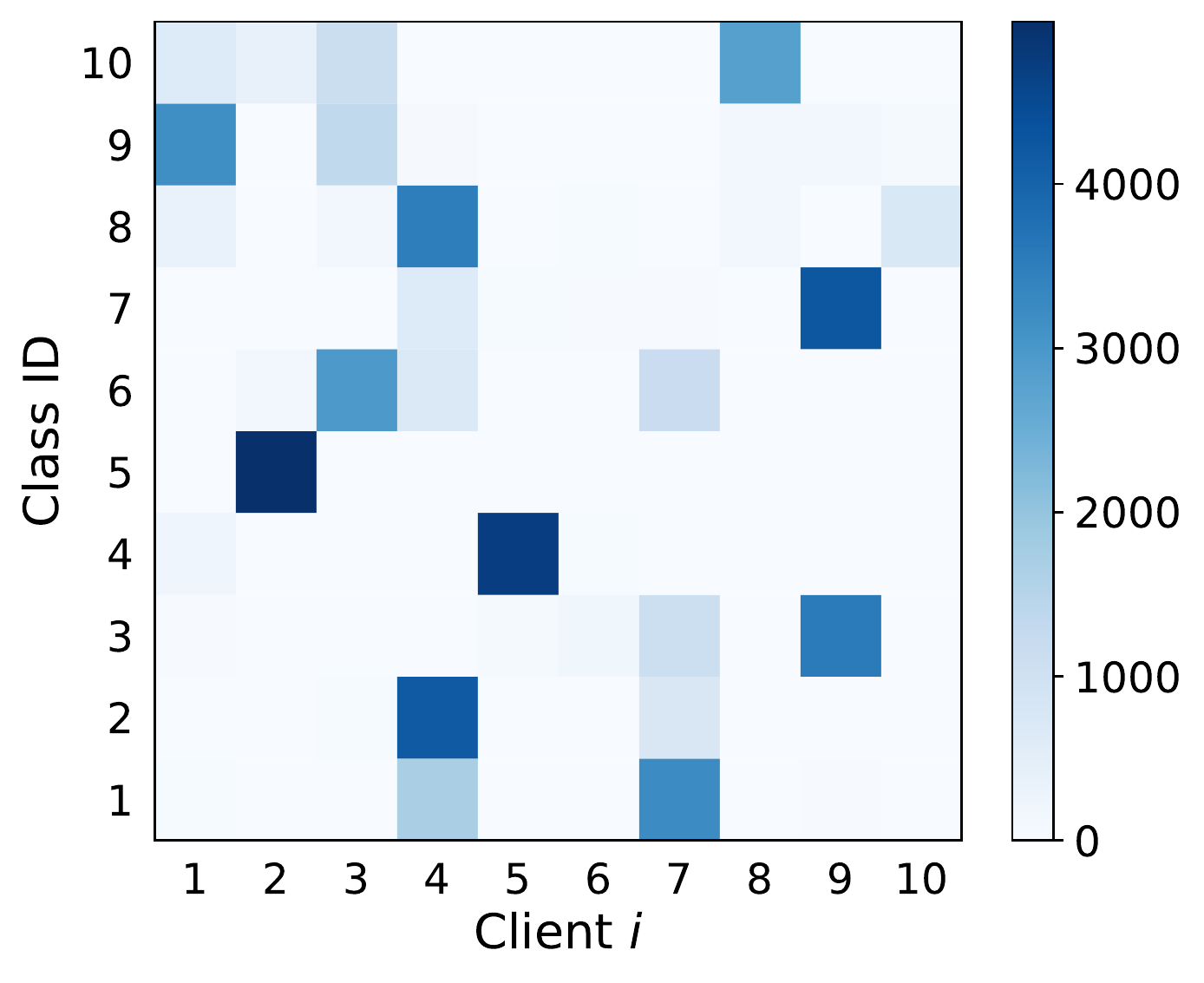} \label{subfig:beta_0.1}} 
	\caption{Client data distribution with different $\beta$ values for the CIFAR-10 dataset. A darker rectangle denotes a higher number of data samples for a specific class in a client.}
	\label{fig:betas}
\end{figure*}

\begin{table*}[tb]
\centering
\caption{Test accuracy (\%) on CIFAR-10 and CIFAR-100 datasets with different $\beta$ values. We tune $\mu$ for each method and the best $\mu$ value is shown in brackets.}
\label{tab:beta_abalation}
\begin{tabular}{l|lll|lll}
\hlineB{3}
        & \multicolumn{3}{c|}{CIFAR-10}                                         & \multicolumn{3}{c}{CIFAR-100}                                     \\
        & $\beta=5$             & $\beta=0.5$           & $\beta=0.1$           & $\beta=5$         & $\beta=0.5$ & $\beta=0.1$                     \\ \hlineB{3}
FedAvg  & 68.69                 & 65.82                 & 55.62                 & \textbf{40.62}    & 40.43                 & 37.48                 \\
FedCKA  & 68.62 (1)             & 66.55 (0.1)           & 55.64 (0.1)           & 40.40 (0.1)       & \textbf{41.91} (10)   & 35.73 (10)            \\
FedProx & 70.03 (0.01)          & 65.03 (0.01)          & 56.55 (0.01)          & 40.00 (0.001)     & 39.91 (0.1)           & 37.38 (0.1)           \\
MOON    & 69.34 (1)             & 66.35 (5)             & 57.60 (1)             & 40.23 (0.1)       & 40.46 (0.1)           & 37.16 (5)             \\ \hline
FedIntR & \textbf{70.66} (5)    & \textbf{67.23} (3)    & \textbf{58.32} (10)   & 40.06 (0.1)       & 41.48 (10)            & \textbf{38.24} (0.5)  \\ \hlineB{3}
\end{tabular}
\end{table*}

To represent the heterogeneous data in the clients, the training portion of each dataset is partitioned into 10 client datasets using the Dirichlet distribution~\cite{ferguson1973bayesian}. The concentration parameter $\beta$ controls the strength of data heterogeneity in the Dirichlet distribution. A lower $\beta$ value indicates a greater level of data heterogeneity. We set the $\beta$ value as 0.5 by default. The performance of FedIntR and the baseline methods is obtained by evaluating the corresponding global models on the testing portions of the respective datasets. We report the median of test accuracy values attained by the global model over the course of last 10 training rounds. Unless otherwise stated, we use the same settings as above for FedIntR and all other baseline methods.

\subsection{Accuracy on Different Datasets}
\label{sec:different_datasets}

Table~\ref{tab:mult_datasets} displays the top-1 test accuracy of different approaches on four datasets. The hyperparameter $\mu$ in FedIntR, MOON~\cite{li2021model}, FedProx~\cite{li2020federated}, and FedCKA~\cite{son2021compare} can be tuned to control the regularization strength in the local training loss. For each approach, we tune the $\mu$ value and report the best results. For FedIntR and FedCKA, we tune $\mu$ from $[0.1, 1, 3, 5, 10]$ (which covers the range used by the original FedCKA paper). We tune $\mu$ from $[0.1, 1, 5, 10]$ for MOON and $[0.001, 0.01, 0.1, 1]$ for FedProx, according to their respective papers. In Table~\ref{tab:mult_datasets}, we report the corresponding best $\mu$ values in parentheses.

Our proposed FedIntR framework achieves superior performance in the SVHN and CIFAR-10 datasets compared to the baseline techniques. For the Fashion-MNIST dataset, both FedIntR and MOON achieve the best performance. FedCKA outperforms FedIntR on the CIFAR-100 dataset, but FedIntR still maintains a comparable performance. There are three plausible explanations for this. (i) For the CIFAR-100 dataset, we employ the ResNet-20 model, which contains fewer number of extraction points for intermediate representations than the CNN model. (ii) The superior performance of FedIntR can be observed more clearly with a higher degree of data heterogeneity, which will be discussed in Section~\ref{sec:data_heterogeneity}, where we compare the performance of different techniques on varying degrees of data heterogeneity. (iii) FedIntR is more beneficial from longer training. By default, we set the number of communication rounds as 100, but as we will discuss in Section~\ref{sec:scalability}, FedIntR performs better with a higher number of training rounds. For reference, we also provide the performance of FedIntR from tuning $\mu$ in Table~\ref{tab:mu_vs_accuracy}.

Note that MOON paper~\cite{li2021model} reported a higher accuracy across CIFAR-10 and CIFAR-100 datasets. As mentioned by the authors of FedCKA~\cite{son2021compare}, MOON may have applied a variety of data augmentations to obtain higher performance. However, since MOON did not disclose their augmentation settings, we were unable to reproduce their results.

\begin{table*}[tbp]
\centering
\caption{Test accuracy (\%) at different rounds $T$ on CIFAR-10 dataset with different number of clients $N$.}
\label{tab:client_abalation}
\begin{tabular}{l|cccc|ccccc}
\hlineB{3}
                  & \multicolumn{4}{c|}{$N=50$}                                       & \multicolumn{5}{c}{$N=100$}                                                        \\
                  & $T=100$        & $T=200$        & $T=500$        & $T=800$        & $T=100$        & $T=200$        & $T=500$        & $T=800$        & $T=1000$       \\ \hlineB{3}
FedAvg            & \textbf{49.41} & 57.79          & 65.26          & 64.62          & 37.98          & \textbf{50.69} & 61.00          & 63.64          & 64.90          \\
FedCKA ($\mu=1$)  & 45.28          & 56.76          & 64.90          & 63.96          & 36.81          & 47.10          & 60.43          & 64.30          & 64.41          \\
FedCKA ($\mu=10$) & 44.45          & 53.83          & 58.69          & 59.08          & 36.68          & 46.85          & 56.51          & 59.91          & 60.49          \\
FedProx ($\mu=0.01$)          & 46.17          & \textbf{58.12} & 65.44          & 65.75          & 37.79          & 50.37          & 61.08          & 64.48          & 64.75          \\
MOON ($\mu=1$)    & 48.25          & 57.98          & 64.24          & 64.23          & \textbf{38.08} & 47.98          & 60.08          & 63.66          & 64.85          \\
MOON ($\mu=10$)   & 42.71          & 55.83          & 64.52          & 65.18          & 34.34          & 46.81          & \textbf{61.68} & 63.97          & 65.12          \\ \hline
FedIntR ($\mu=1$)   & 47.02          & 58.01          & 63.89          & 63.75          & 35.87          & 48.15          & 60.59          & 63.46          & 64.26          \\
FedIntR ($\mu=10$)  & 46.63          & 56.39          & \textbf{65.48} & \textbf{66.94} & 32.13          & 49.76          & 61.31          & \textbf{65.55} & \textbf{66.57} \\ \hlineB{3}
\end{tabular}
\end{table*}

\subsection{Degree of Data Heterogeneity}
\label{sec:data_heterogeneity}

In this experiment, we investigate the impact of clients’ data heterogeneity on our proposed FedIntR framework. We adjust the $\beta$ parameter of the Dirichlet distribution to control the strength of data heterogeneity between the clients. Fig.~\ref{fig:betas} illustrates how different values of $\beta$ influence the local data distributions of the clients for the CIFAR-10 dataset.

Table~\ref{tab:beta_abalation} displays the test accuracy values attained by different approaches on the CIFAR-10 and CIFAR-100 datasets with $\beta$ values of 5, 0.5, and 0.1. Similar to Section~\ref{sec:different_datasets}, we tune the $\mu$ value for each approach (shown in parentheses in Table~\ref{tab:beta_abalation}) and report the best results. For the CIFAR-10 dataset, FedIntR exhibits superior performance over the baseline approaches in all data heterogeneity settings. FedProx has comparable performance to FedIntR when $\beta=5$ (i.e., when the degree of data heterogeneity among the clients is low), but its performance diminishes with a smaller $\beta$ value. For the CIFAR-100 dataset, vanilla FedAvg outperforms all other approaches when $\beta=5$. However, as the $\beta$ value decreases, incorporating a form of regularization into the local loss can outperform the vanilla FedAvg. With $\beta=0.1$, our proposed FedIntR significantly outperforms the other approaches. We speculate that fine-grained regularization computed from intermediate representations and layer-wise contributions makes FedIntR more robust in FL environments with high degrees of data heterogeneity.

\subsection{Scalability}
\label{sec:scalability}

To ensure that FedIntR is scalable in FL environments with a large number of clients, we conduct experiments on the CIFAR-10 dataset with 50 and 100 client settings (i.e., $N=50$ and $N=100$). For both settings, we use the Dirichlet distribution with $\beta=0.5$ to partition the data, and 20\% of the clients are chosen at random to participate in each training round. In particular, we uniformly sample 10 clients for $N=50$ and 20 clients for $N=100$ settings. For FedProx, $\mu$ value is set to 0.01, and for all other approaches, we experiment with both $\mu=1$ and $\mu=10$. We set the number of training rounds $T$ as 800 for $N=50$ and 1000 for $N=100$. 

\begin{figure}[tbp]
    \centering
    \includegraphics[width=0.98\linewidth]{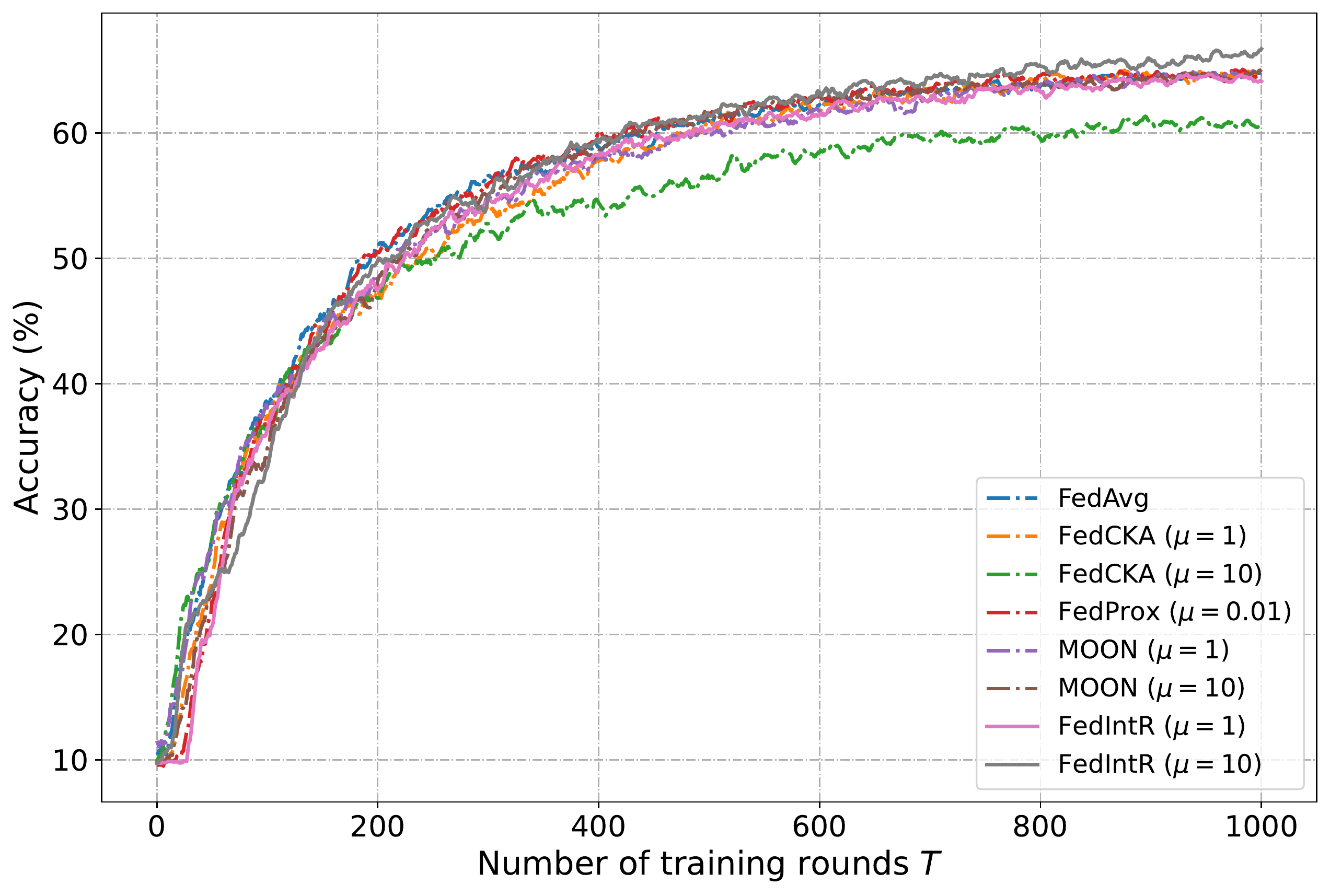}
    \caption{Test accuracy (\%) on CIFAR-10 dataset with number of clients $N=100$. }
    \label{fig:scalability}
\end{figure}

From Table~\ref{tab:client_abalation}, we can observe that vanilla FedAvg outperforms the majority of other approaches in the early training rounds (i.e., $T=100$ and $T=200$) for both $N=50$ and $N=100$ settings. This is due to the fact that the local training loss in FedAvg focuses only on performance improvement, whereas the other approaches include a regularization term with additional objectives. However, with a sufficient number of training rounds, FedIntR with $\mu=10$ can significantly outperform all other approaches. In the $N=50$ setting, FedIntR ($\mu=10$) outperforms the second best approach, FedProx, by 1.19\% at $T=800$, and in the $N=100$ setting, it outperforms MOON ($\mu=10$) by 1.45\% at $T=1000$. Fig.~\ref{fig:scalability} shows the test accuracy values of different methods in each round for the $N=100$ setting.

\subsection{Number of Local Epochs}
\label{sec:local_epochs}

\begin{figure}[tbp]
    \centering
    \includegraphics[width=0.98\linewidth]{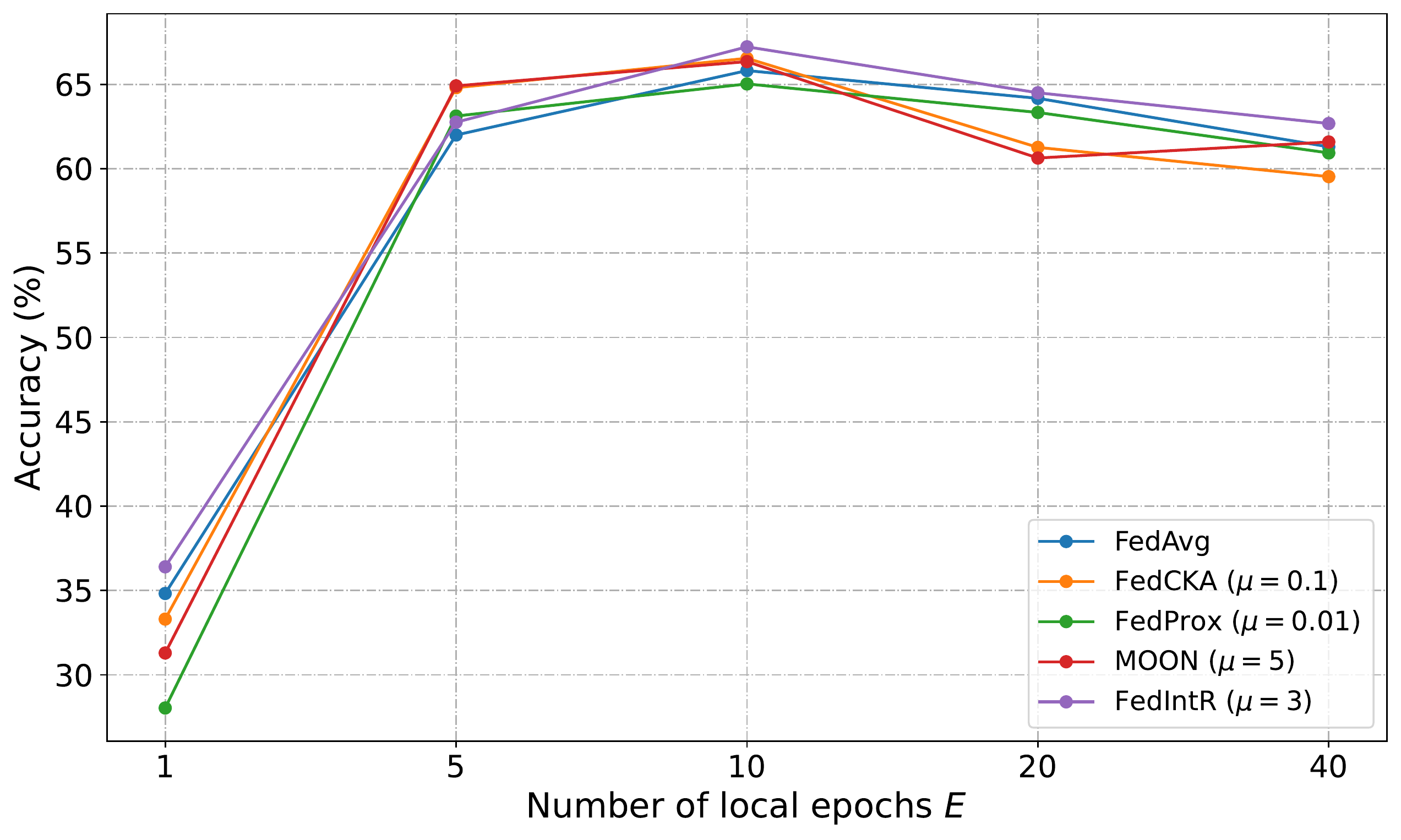}
    \caption{Test accuracy (\%) with different numbers of local epochs on the CIFAR-10 dataset. For each approach, we use the same $\mu$ values tuned with the CIFAR-10 dataset as in Table~\ref{tab:mult_datasets}.}
    \label{fig:le_abalation}
\end{figure}

Using the CIFAR-10 dataset, we also explore the effects of the number of local epochs $E$ on different approaches. The experiments are performed using 1, 5, 10 (default), 20, and 40 local epochs with 100 rounds of training. For each approach, we use the same $\mu$ values tuned with the CIFAR-10 dataset in Section~\ref{sec:different_datasets} and the results are shown in Fig.~\ref{fig:le_abalation}. Consistent with the findings in MOON, setting $E=1$ results in poor accuracy for all approaches, while setting $E=10$ enables them to achieve their best performance. Due to the drift towards the local data distribution, using a larger number of local epochs (i.e., 20 or 40) can decrease the performance for all approaches. Nonetheless, our proposed method, FedIntR, achieves the best performance in most settings except $E=5$.

\subsection{Ablation on Layer-wise Weights}

FedIntR makes use of the layer-wise weight $\alpha_k$ that determines the contribution of each individual representation loss $\ell_k$ to the regularization term. To ensure that $\alpha_k$ is a crucial component of FedIntR for improving the performance, we conduct an ablation study by computing the regularization term in an alternative way. Instead of incorporating $\alpha_k$ into the regularization term, we use the average of $[\ell_k]_{k=1}^{K}$ as the regularization term. Specifically, we apply the local loss function defined as follows:
\begin{equation}
    \mathcal{L} = \ell_{sup} + \frac{\mu}{K} \sum^K_{k=1} \ell_k
    \label{eqn:avg_reg_local_loss}.
\end{equation}

In Table~\ref{tab:ir_weights_abalation}, we compare the performance of FedIntR with and without $\alpha_k$ on four datasets. We can observe that calculating a layer-wise weight $\alpha_k$ for its corresponding representation loss $\ell_k$ can significantly improve the performance of the resulting global model. $\alpha_k$ determines the contribution of individual $\ell_k$ to the regularization term, enabling FedIntR to perform local training with fine-grained regularization.

\begin{table}[tbp]
\centering
\caption{Test accuracy (\%) of FedIntR with and without $\alpha_i$ for different datasets. We use the same $\mu$ values tuned for each dataset as in Table \ref{tab:mult_datasets}.}
\label{tab:ir_weights_abalation}
\begin{tabular}{l|llll}
\hlineB{3}
           & \begin{tabular}[c]{@{}l@{}}Fashion-\\ MNIST\end{tabular}   & SVHN              & CIFAR-10 & CIFAR-100                  \\ \hlineB{3}
$\alpha_i$ & \textbf{89.15} & \textbf{87.17}    & \textbf{67.23}    & \textbf{41.48}    \\
Average    & 89.03          & 85.62             & 64.20             & 40.33             \\ \hlineB{3}
\end{tabular}
\end{table}

\section{Conclusion}

Nowadays, data valuable for powering intelligent services may be distributed over numerous personal edge devices. This renders data collection impractical due to privacy concerns, and hence the role of federated learning substantially grows. Client data heterogeneity in FL is one major shortcoming that is responsible for the poor performance of global model. To mitigate this, we propose FedIntR, which introduces a simple and effective regularization term to complement the local training process of FL. FedIntR uses intermediate layer representations to incorporate more information into the regularization process. Moreover, the layer-wise weights in FedIntR automatically determine each representation's contribution to the regularization term. This enables FedIntR to produce fine-grained regularization for the local training process, thus enhancing the global model performance. We conduct extensive experiments on various datasets and FL settings to examine the efficacy of FedIntR. Comparison between FedIntR and various state-of-the-art methods demonstrates that FedIntR achieves superior performance in most settings.

\bibliographystyle{IEEEtran}
\bibliography{main}

\end{document}